\documentclass{bmvc2k}
\usepackage{booktabs}
\usepackage{multirow}
\usepackage{amssymb}
\usepackage{amsmath}
\usepackage{amsfonts}
\usepackage{cleveref}
\usepackage{float}
\title{FaceCrafter: Identity-Conditional Diffusion with Disentangled Control over Facial Pose, Expression, and Emotion}
\addauthor{Kazuaki Mishima}{mishima.k.ad@m.titech.ac.jp}{1}
\addauthor{Antoni Bigata Casademunt}{a.bigata-casademunt22@imperial.ac.uk}{2}
\addauthor{Stavros Petridis}{stavros.petridis04@imperial.ac.uk}{2}
\addauthor{Maja Pantic}{m.pantic@imperial.ac.uk}{2}
\addauthor{Kenji Suzuki}{suzuki.k.di@m.titech.ac.jp}{1}

\addinstitution{
 Institute of Science Tokyo\\
 Tokyo, Japan
}

\addinstitution{
 Imperial College London\\
 London, United Kingdom
}

\runninghead{Mishima, et al.}{FaceCrafter: ID-Conditional Diffusion}

\begin{document}

\maketitle

\begin{abstract}

Human facial images encode a rich spectrum of information, encompassing both stable identity-related traits and mutable attributes such as pose, expression, and emotion.
While recent advances in image generation have enabled high-quality identity-conditional face synthesis, precise control over non-identity attributes remains challenging, and disentangling identity from these mutable factors is particularly difficult.
To address these limitations, we propose a novel identity-conditional diffusion model that introduces two lightweight control modules designed to independently manipulate facial pose, expression, and emotion without compromising identity preservation.
These modules are embedded within the cross-attention layers of the base diffusion model, enabling precise attribute control with minimal parameter overhead.
Furthermore, our tailored training strategy, which leverages cross-attention between the identity feature and each non-identity control feature, encourages identity features to remain orthogonal to control signals, enhancing controllability and diversity.
Quantitative and qualitative evaluations, along with perceptual user studies, demonstrate that our method surpasses existing approaches in terms of control accuracy over pose, expression, and emotion, while also improving generative diversity under identity-only conditioning. Our project page is \href{https://mishimario.github.io/FaceCrafter/}{here}.

\end{abstract}

%-------------------------------------------------------------------------
\section{Introduction}

Human facial images convey rich and multifaceted information that can be broadly categorized into stable identity-related features---such as facial bone structure, gender, and age---and mutable attributes, including pose, expression, emotion, hairstyle, and accessories (e.g., glasses). An ideal facial image dataset should encompass a diverse range of both identity and mutable attributes, as such diversity is crucial for advancing facial image analysis and generation research.

However, existing large-scale facial datasets~\cite{huang2008labeled_lfw_dataset, liu2015deep_celeb_dataset, karras2017progressive_celeb_hq, mollahosseini2017affectnet, li2017reliable_raf-db,karras2019style_FFHQ} used in recognition or generation research often suffer from severe imbalances, particularly in pose, expression, and emotional diversity. For example, AffectNet~\cite{mollahosseini2017affectnet}, a widely used benchmark for facial affect recognition, contains approximately 70\% of samples labeled as either \textit{Happy} or \textit{Neutral}, limiting the variability required for robust model training and evaluation.

Meanwhile, recent advances in generative models---such as GANs~\cite{goodfellow2014generative_GAN,karras2019style_stylegan, karras2020analyzing_stylegan2, karras2021alias_stylegan3} and diffusion models~\cite{ddpm}---have dramatically improved the fidelity of synthesized facial images. Yet, simultaneously preserving identity while enabling fine-grained control over mutable attributes remains a challenging task. Although prior works~\cite{huang2017beyond_pose_gan, hu2018pose_gan, ding2018exprgan_ExprGan_expedit, lindt2019facial_expedit,pumarola2018ganimation_expedit, choi2018stargan_expedit, tang2022facial_emo_gan, shen2020interpreting_InterFaceGAN, patashnik2021styleclip, xia2021tedigan_gan_face_control, drobyshev2024emoportraits, liang2024caphuman} on face editing have explored controlling pose, expression, or emotion, few studies have attempted to simultaneously control all three attributes.
Moreover, these approaches focus mainly on control itself and do not explicitly pursue the disentanglement between identity and non-identity (mutable) factors.
Furthermore, while frameworks like ControlNet~\cite{zhang2023adding_controlnet} have shown that controllable generation via additional input signals is feasible, such methods introduce substantial parameter overhead. 

Arc2Face~\cite{papantoniou2024arc2face}, a pioneering identity-conditional diffusion foundation model, can synthesize diverse faces of a single identity from a single ID image and serves as a versatile base model for various attribute control and synthetic dataset generation applications. However, Arc2Face does not explicitly disentangle identity from attributes like pose, expression, or emotion. As a result, its entangled representations hinder precise control and degrade the purity of learned identity features.

% To overcome these limitations, we propose a novel identity-conditional diffusion framework that achieves precise and independent control over facial pose, expression, and emotion while ensuring strong identity preservation and maintaining low computational overhead. Furthermore, our method realizes a more disentangled identity-conditional foundation model, facilitating clearer separation between identity and non-identity representations.

% Our approach introduces two lightweight control modules---one for pose and expression and another for emotion---which are embedded within the cross-attention mechanism of the base diffusion model. Additionally, we propose a tailored learning strategy that explicitly encourages identity features to remain independent from non-identity control features, thereby enhancing controllability, and improving diversity without relying on non-identity control.

% To overcome these limitations, we propose a lightweight identity-conditional diffusion framework that enables precise and independent control over facial pose, expression, and emotion, while preserving identity and maintaining low computational cost. Our approach introduces two control modules—one for pose/expression and one for emotion into the cross-attention layers of the diffusion model. Combined with a tailored learning strategy that explicitly disentangles identity from non-identity features, our method enhances controllability, improves diversity, and ensures robust identity preservation.

To overcome these limitations, we propose FaceCrafter, a lightweight identity-conditional diffusion framework that enables precise and independent control over facial pose, expression, and emotion, while preserving identity. Our approach introduces two control modules—one for pose/expression and one for emotion—into the cross-attention layers of the diffusion model. Combined with a tailored learning strategy that explicitly disentangles identity from non-identity features, FaceCrafter enhances controllability, improves diversity, and ensures robust identity preservation.

Our contributions are summarized as follows:
\begin{itemize}
    \item We introduce two lightweight control modules, one for pose/expression and one for emotion, embedded within the cross-attention layers of the base diffusion model.
    \item We propose a tailored learning strategy that enforces orthogonality between identity features and non-identity mutable control features, enhancing both controllability and generative diversity.
    % \item We conduct \textbf{comprehensive quantitative and qualitative experiments}, including perceptual user studies, demonstrating superior control accuracy and generative diversity compared to state-of-the-art methods.
  \item Our model controls pose, expression, and emotion more accurately with fewer parameters and surpasses the current state-of-the-art ID-conditional foundation model, Arc2Face in diversity and realism using only ID condition.
  % Our model achieves superior accuracy in controlling facial pose, expression, and emotion with relatively few additional parameters compared to state-of-the-art methods. Furthermore, it outperforms the current state-of-the-art ID-conditional foundation model, Arc2Face, in both generative diversity and realism, with only ID condition.
\end{itemize}

\section{Related Works}
\textbf{Face Generative Models.} Recent advances in face image generation have been largely driven by the development of StyleGAN~\cite{karras2019style_stylegan, karras2020analyzing_stylegan2, karras2021alias_stylegan3}. Various
methods have been proposed based on it, including approaches that enable image editing in the latent space,
text-guided manipulation, and multimodal control, providing flexible and diverse ways of controlling image generation~\cite{shen2020interpreting_InterFaceGAN, harkonen2020ganspace, 
nitzan2022mystyle_personalized_generative_prior,patashnik2021styleclip, xia2021tedigan_gan_face_control}. 
Following that, diffusion models~\cite{ddpm} have rapidly progressed and demonstrated superior image quality 
compared to GANs, leading to increasing attention in face image generation based on diffusion models. 
% Particularly 
% in ID-conditional generation, several approaches have been proposed, including methods that personalize models with
% fine-tuning during inference~\cite{gal2022image_textual_inversion, ruiz2023dreambooth, ruiz2024hyperdreambooth, gal2023encoder_personalized, zhou2023enhancing_profusion_personalized} and those that incorporate additional modules to explicitly learn ID-related features~\cite{xiao2024fastcomposer, li2024photomaker, zhang2024flashface, chen2023photoverse_similar_ip-adapter, valevski2023face0, chen2024dreamidentity, peng2024portraitbooth, wang2024instantid_big_model, yan2023facestudio_similar_ip-adapter, he2024imagine_yourself_meta_big_model}.
Particularly in ID-conditional generation, existing approaches can be broadly categorized into methods that require inference-time fine-tuning~\cite{gal2022image_textual_inversion, ruiz2023dreambooth, ruiz2024hyperdreambooth, gal2023encoder_personalized, zhou2023enhancing_profusion_personalized}and those that leverage additional modules to encode ID features without inference-time fine-tuning~\cite{xiao2024fastcomposer, li2024photomaker, zhang2024flashface, chen2023photoverse_similar_ip-adapter, valevski2023face0, chen2024dreamidentity, peng2024portraitbooth, wang2024instantid_big_model, yan2023facestudio_similar_ip-adapter, he2024imagine_yourself_meta_big_model}.
% Moreover, it has become possible to control via not only ID input and text but also facial expressions and body 
% movements using multiple conditions, which has fostered research on multi-conditional generation. 

\noindent
\textbf{Diffusion Control Methods.} Among conditioning methods, ControlNet~\cite{zhang2023adding_controlnet} 
is widely used due to its ability to smoothly and precisely guide the generation process, and several extensions have been proposed~\cite{zhao2023uni-controlnet, bansal2023universal_diffusion}. However, ControlNet comes at the price of a large number of additional
parameters. As an alternative, IP-Adapter~\cite{ye2023ip-adapter} has been introduced to control generation using
condition extractors and minimal additional parameters in cross-attention layers. Several methods adopting similar
control structures have also been developed~\cite{chen2023photoverse_similar_ip-adapter, liu2024ada_similar-ip-adapter, he2024imagine_yourself_meta_big_model}. 

\noindent
\textbf{Attribute Control on Face Generation.} 
% ID-conditional generation often involves synthesizing human faces while preserving the given identity. 
% Since text alone typically does not capture ID-specific information, many studies have explored combining ID image condition with text-image generation.
Beyond ID conditioning, approaches for controlling pose, expression, and complex emotions have been proposed~\cite{huang2017beyond_pose_gan, hu2018pose_gan, ding2018exprgan_ExprGan_expedit, lindt2019facial_expedit,pumarola2018ganimation_expedit, choi2018stargan_expedit, tang2022facial_emo_gan, shen2020interpreting_InterFaceGAN, patashnik2021styleclip, xia2021tedigan_gan_face_control, drobyshev2024emoportraits, azari2024emostyle, liang2024caphuman}. For example, EmoPortraits~\cite{drobyshev2024emoportraits} extracts pose and emotion features from a target face and combines them to enable fine-grained face reenactment. EmoStyle~\cite{azari2024emostyle}, based on StyleGAN2, controls emotional expression smoothly using two dimensions of emotion: valence and arousal~\cite{russell1980circumplex_Valence_Arousal}. Furthermore, CapHuman~\cite{liang2024caphuman} employs a ControlNet-like structure to enable generation conditioned on text, ID, pose, and expression simultaneously.

\noindent
\textbf{ID-conditional Foundation Model.}
While many prior diffusion-based studies leverage general-purpose models such as Stable Diffusion~\cite{rombach2022high_stable_diffusion} without additional full-fine-tuning~\cite{zhang2024flashface,chen2023photoverse_similar_ip-adapter,chen2024dreamidentity,he2024imagine_yourself_meta_big_model,ye2023ip-adapter}, a recent work, Arc2Face~\cite{papantoniou2024arc2face}, introduces an ID-conditional foundation model specifically tailored for face generation. Arc2Face embeds ArcFace~\cite{deng2019arcface} features as tokens and inputs them into a CLIP~\cite{radford2021learning_CLIP} text encoder to create ID embeddings, which are then used to fine-tune a base model with a large-scale dataset of approximately 42 million face images. This approach enables diverse face generation and facilitates research in face synthesis datasets and ID-conditional control of pose and expression beyond mere image editing.
Building on Arc2Face, we propose a lightweight model that simultaneously controls pose, expression, and emotion, enabling natural generation with preserved target emotions. By disentangling identity from non-ID features, our method achieves more robust ID-conditional generation.
% Building upon Arc2Face, our work proposes a lightweight architecture capable of controlling pose, expression, and emotion simultaneously. While many prior works address pose, expression, and emotion control separately, we integrate emotion control to enable more natural generation while preserving target emotions. Furthermore, by explicitly disentangling intrinsic ID features from non-ID features such as pose, expression, and emotion, we realize a more robust ID-conditional model. 

\begin{figure}[htb]
    \centering
    \includegraphics[width=1.0\linewidth]{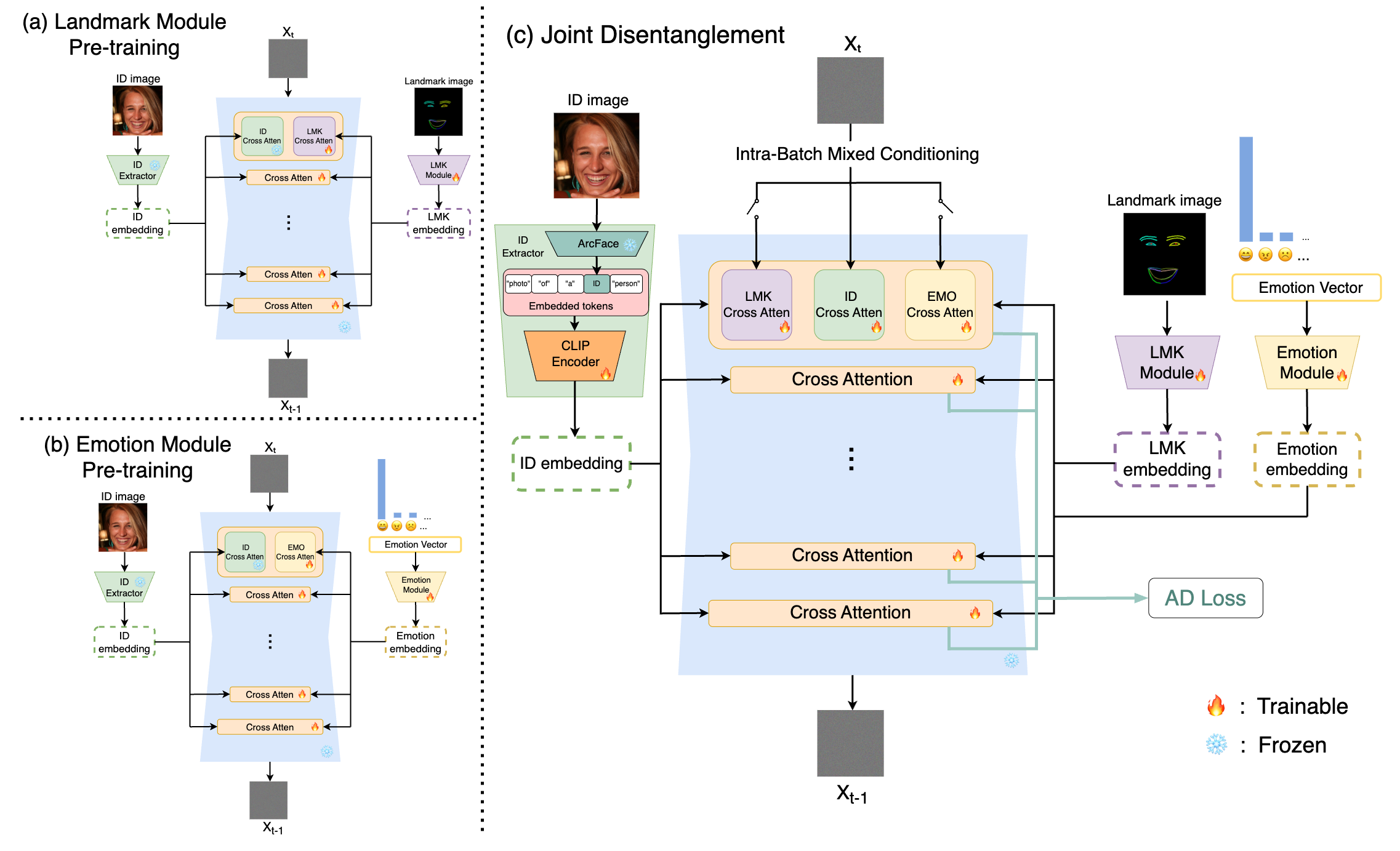}
    \caption{Overview of FaceCrafter. (a) and (b) illustrate the Pre-training schemes of the control modules using landmark and emotion vector, respectively. (c) presents the Joint Disentanglement training framework that combines both modules. During training, only the control modules and their corresponding cross-attention layers are trained, while all other parts are frozen. In (c), we introduce Intra-Batch Mixed Conditioning, where different combinations of conditions are mixed within a batch, along with Attention Disentanglement (AD) Loss to increase the distance between feature representations of different conditions.}
    \label{fig:overview}
\end{figure}

\section{Our Approach}

% We provide an overview of our proposed method in Figure~\ref{fig:overview}. 
An overview of our proposed method, FaceCrafter, is shown in Figure~\ref{fig:overview}.
% In the pre-training phase, our approach separately learns to control facial pose and expression from landmark representations, and emotion from an emotion vector. Then, we perform an overall training phase in which both types of control are jointly optimized. During this phase, we introduce an Attention Disentanglement Loss (AD Loss) that encourages the features between cross-attention layers under mixed conditions to be separated, thereby promoting disentanglement between identity and non-identity attributes.
Sections~\ref{sec: method_model_arch}, ~\ref{sec: method_training_framework}, and ~\ref{sec: method_infer} describe the model architecture, the training framework, and the inference procedure during the generation stage, respectively.
For clarity and brevity, we denote identity, landmark, and emotion conditions as \textbf{ID}, \textbf{LMK}, and \textbf{EMO}, respectively.

% \begin{figure}[htb]
%     \centering
%     % \includegraphics[width=1.0\linewidth]
%     \includegraphics[width=8.0cm, height=5.0cm]{images/ADLoss_overview.pdf}
%     \caption{This figure illustrates the overview of ADLoss.
% It is designed to increase the separation between the attention maps, thereby promoting better disentanglement of ID, LMK, and EMO features.}
%     \label{fig:ADLoss_overview}
% \end{figure}

\subsection{Model Architecture}
\label{sec: method_model_arch}
\textbf{Base Model.} 
% In this study, we use Arc2Face~\cite{papantoniou2024arc2face}, first introduced as the ID-conditional foundation model, as our base model. Specifically, it embeds features from the face recognition model ArcFace~\cite{deng2019arcface} as a word token in the CLIP~\cite{radford2021learning_CLIP} text encoder, and this embedding is used as the ID condition to train a model based on Stable Diffusion~\cite{rombach2022high_stable_diffusion}.
We adopt Arc2Face~\cite{papantoniou2024arc2face} as our base model, which serves as an identity-conditional generation framework. It incorporates features extracted from ArcFace~\cite{deng2019arcface} as word tokens into the CLIP~\cite{radford2021learning_CLIP} text encoder, and uses these embeddings as identity conditions to train a model based on Stable Diffusion~\cite{rombach2022high_stable_diffusion}.

\noindent
\textbf{Landmark/Emotion Control Modules.} We control pose and expression using facial landmark images generated by MediaPipe~\cite{lugaresi2019mediapipe}, processed by a CNN-based Landmark Control Module that encodes and flattens them into a sequence. To additionally handle emotions, we introduce an Emotion Control Module that takes Valence, Arousal~\cite{russell1980circumplex_Valence_Arousal}, and 8-class emotion distributions from EmoNet~\cite{gerczuk2021emonet} as input, processes them via a DNN, and reshapes the output into a sequence.
Details of each module are provided in the supplementary material.

\noindent
\textbf{Cross Attention Summation.} Following IP-Adapter~\cite{ye2023ip-adapter}, we combine multiple cross-attention modules via summation within the attention block. Given input feature $\mathbf{x}$ and condition features $\mathbf{c}$, cross-attention is computed as $\mathrm{Attention}(\mathbf{Q}, \mathbf{K}, \mathbf{V}) = \mathrm{softmax}(\mathbf{QK}^\top / \sqrt{d}) \mathbf{V}$, where $\mathbf{Q} = \mathbf{x}W_q$, $\mathbf{K} = \mathbf{c}W_k$, and $\mathbf{V} = \mathbf{c}W_v$. We use three conditions, identity, landmark, and emotion, with final attention output: 
\begin{equation}
\begin{aligned}
\mathrm{Attention}_{\text{final}} = \mathrm{Attn}_{\text{ID}} + \alpha_{\text{LMK}} \mathrm{Attn}_{\text{LMK}} + \alpha_{\text{EMO}} \mathrm{Attn}_{\text{EMO}}
\end{aligned}
\end{equation}
% where $\alpha$ are binary weights that enable/disable each branch. All parameters are trained jointly.
The coefficients $\alpha_{\text{LMK}}$ and $\alpha_{\text{EMO}}$ 
serve as control weights for the respective attention branches. 
In our experiments, these values are binary (0 or 1), acting effectively as switches to enable or disable the use of each condition. 
% All learnable parameters, including those in the key and value projections for each condition, are trained jointly as part of the overall model.

\subsection{Training Framework}
\label{sec: method_training_framework}
% \textbf{Collecting images for our training.} For training, a diverse set of face image datasets is necessary. The widely used FFHQ~\cite{karras2019style_FFHQ} dataset includes face images from various ethnicities, genders, and ages, making it a high-quality image set in terms of ID. However, it is not diverse in terms of facial poses, expressions, and emotions, making it insufficient for control-based learning in these aspects. To address this limitation, a more diverse dataset comprising different poses, expressions, and emotions is constructed by cropping face images from video datasets of FEED~\cite{drobyshev2024emoportraits} and MEAD~\cite{kaisiyuan2020mead} dataset, and combining them with the FFHQ dataset. This enabled the creation of a dataset that is not only diverse in terms of ID but also in terms of pose, expression, and emotion.

\textbf{Pre-training Stage.} To ensurre a stable training we train each module independently, as illustrated in Figure~\ref{fig:overview} (a) and (b). When using datasets that contain multiple poses and expressions per identity, we adopt a sampling strategy where, for each pair, the identity image and the input image are selected as different images of the same identity. Specifically, when training the Landmark Control Module, the input image and the ID image are randomly selected to have different poses and facial expressions. In contrast, when training the Emotion Control Module, images with different facial expressions are sampled while keeping the pose fixed. In this stage, only each control module and its corresponding key and value projection layers are trained, while the rest of the network remains frozen.
% The overall training phase is conducted using the individually pre-trained control modules.

\noindent
\textbf{Joint Disentanglement Stage.} 
% In Stage 2, the ID extractor part is combined with the LMK control module and the Emotion control module for joint training. To further separate the features of the ID extractor and the LMK/Emotion modules, an Attention Decouple Loss is introduced. Specifically, for each attention feature in the respective layers, the attention values are computed as follows:
In this stage, we jointly train the entire control framework by integrating two pre-trained control modules and the ID extractor. An overview of this framework is illustrated in Figure~\ref{fig:overview} (c). To facilitate the disentanglement of different control conditions including ID, we propose a novel training strategy that explicitly separates the cross-attention features associated with different combinations of conditions.
We first replicate each sampled data point $B$ times, where $B$ is the batch size. For each replicated sample within the batch, we assign one of the following conditioning combinations with equal probability: (1) ID only, (2) ID + LMK, (3) ID + EMO, and (4) ID + LMK + EMO. This is implemented by assigning attention scaling factors $(\alpha_\text{LMK}, \alpha_\text{EMO})$ as $(0.0, 0.0)$, $(1.0, 0.0)$, $(0.0, 1.0)$, and $(1.0, 1.0)$, respectively. We refer to this strategy as \textbf{Intra-Batch Mixed Conditioning}.
Based on this conditioning setup, we propose a novel loss function called the \textbf{Attention Disentanglement Loss} (\(\mathcal{L}_\text{AD}\)) to explicitly encourage the separation of attention features associated with different control combinations. This loss penalizes similarity in attention representations between samples that differ in conditioning signals, thereby promoting decoupled and interpretable control representations. We define the disentanglement loss between two different conditioning setups \( a \) and \( b \) as follows:

% \frac{1}{|\mathcal{C}^{(a,b)}_n|} 
% \begin{equation}
% \mathcal{L}_{\text{consistency}}(a,b) = 
% \frac{1}{N} \sum_{n=1}^{N} 
% M_{n} \odot \sum_{(i,j) \in \mathcal{C}^{(a,b)}_n} 
% \cos\left( A^{(n)}_{i,a}, A^{(n)}_{j,b} \right)
% \end{equation}

% \begin{equation}
% \mathcal{L}_{\text{decoupled}}(a,b) = 
% \sum_{(i,j) \in \mathcal{C}_n^{(a,b)}} \mathbb{E}_{n \in N} \left[\mathbb{E}_{s \in S_n} \left[ M_{n} \cdot \cos \left( A^{(n)}_{i,a}, A^{(n)}_{j,b} \right) \right]\right]
% \end{equation}

% \begin{equation}
% \mathcal{L}_{\text{decoupled}}(a,b) = 
% \mathbb{E}_{n \in N} \left[ \mathbb{E}_{s \in S_n} \left[ \sum_{(i,j) \in \mathcal{C}_n^{(a,b)}} M_{n} \cdot \cos \left( A^{(n)}_{i,a}, A^{(n)}_{j,b} \right) \right] \right]
% \end{equation}

% \begin{equation}
% \mathcal{L}_{\text{decoupled}}(a,b) = 
% \mathbb{E}_{n \in N} \left[ \mathbb{E}_{s \in S_n} \left[ \sum_{(i,j) \in \mathcal{C}_n^{(a,b)}} M_{n} \cdot \operatorname{ReLU} \left( \cos \left( A^{(n)}_{i,a}, A^{(n)}_{j,b} \right) \right) \right] \right]
% \end{equation}

\begin{equation}
\mathcal{L}_{\text{disentanglement}}(a,b) = 
\mathbb{E}_{n \in N} \left[  \sum_{(i,j) \in \mathcal{C}_n^{(a,b)}} M_{n} \odot \operatorname{ReLU} \left( \cos \left( A^{(n)}_{i,a}, A^{(n)}_{j,b} \right) \right) \right] 
\end{equation}
% \begin{equation}
% \mathcal{L}_{\text{consistency}}(a,b) = 
% \frac{1}{N} \sum_{n=1}^{N} 
% \frac{1}{S} \sum_{(i,j) \in \mathcal{C}^{(a,b)}_n} 
% \sum_{s}^{S}M_{n,s}*\cos\left( A^{(n)}_{i,s,a}, A^{(n)}_{j,s,b} \right)
% \end{equation}

\noindent
Here, $N$ is the number of cross-attention blocks, and $\mathcal{C}^{(a,b)}_n$ is the index set for condition pairs $a,b$ (ID, LMK, EMO) in block $n$. $M_n$ is a binary face mask from a segmentation model~\cite{luo2020ehanet}, resized to match the sequence length. $A^{(n)}_{i,a}$ is the attention feature of sample $i$ at block $n$ for condition $a$, and $\cos(\cdot,\cdot)$ denotes cosine similarity. Using these, the Attention Disentanglement Loss $\mathcal{L}_{\text{AD}}$ is formulated as:

% Here, $N$ denotes the number of cross-attention blocks in the model. For each block $n$, $\mathcal{C}^{(a,b)}_n$ represents the set of index pairs $(i,j)$ that correspond to the different conditions $a$ and $b$ within that block. The conditions $a$ and $b$ can refer to different control modules, such as ID, Landmark (LMK), or Emotion (EMO). 
% The mask \( M_n \) is a binary mask derived from an external pre-trained segmentation model~\cite{luo2020ehanet}, where only facial regions are retained. This mask is resized to match the sequence length to ensure proper alignment during training. $A^{(n)}_{i,a}$ represents the attention feature from the $i$-th sample in the batch at cross-attention block $n$ for input type $a$. $\cos(\cdot, \cdot)$ computes the cosine similarity.
% Using the above definition, the Attention Disentanglement Loss \(\mathcal{L}_{\text{AD}}\) is formulated as:
% Here, $\mathcal{C}_n$ is the set of all valid pairs in batch $n$ such that $(a, b)$ corresponds to the same identity but different combinations of control conditions (e.g., $(\text{ID}, \text{ID+LMK})$, $(\text{ID}, \text{ID+EMO})$, etc.), if both exist in the batch.
\begin{equation}
\begin{aligned}
\mathcal{L}_{\text{AD}} = 
\mathcal{L}_{\text{disentanglement}}&(\text{ID}, \text{ID+LMK}) + 
\mathcal{L}_{\text{disentanglement}}(\text{ID}, \text{ID+EMO}) +  \\
\mathcal{L}_{\text{disentanglement}}&(\text{ID}, \text{ID+LMK+EMO}) + 
\mathcal{L}_{\text{disentanglement}}(\text{ID+LMK}, \text{ID+EMO}) + \\
\mathcal{L}_{\text{disentanglement}}&(\text{ID+LMK}, \text{ID+LMK+EMO})
\end{aligned}
\end{equation}

% An overview of this loss formulation is illustrated in Figure~\ref{fig:ADLoss_overview}. 
\noindent
Therefore, the final training objective combines the conventional diffusion loss with our proposed Attention Disentanglement (AD) Loss, and is defined as:
% \begin{equation}
% \mathcal{L}_{\text{diffusion}} = \mathbb{E}_{x_0, \epsilon, t} \left[ 
% \left\| \epsilon - \epsilon_\theta(x_t, t) \right\|_2^2
% \right]
% \end{equation}

\begin{equation}
\mathcal{L}_{\text{diffusion}} = \mathbb{E}_{x_0, \epsilon, t, c_{\text{ID}}, c_{\text{LMK}}, c_{\text{EMO}}} \left[ 
\left\| \epsilon - \epsilon_\theta(x_t, t, c_{\text{ID}}, c_{\text{LMK}}, c_{\text{EMO}}) \right\|_2^2
\right]
\end{equation}

% \begin{equation}
% \mathcal{L}_{\text{focus}} = 
% \mathbb{E}_{x_0, \epsilon, t} \left[ 
% \left\| M \odot \left( \epsilon - \epsilon_\theta(x_t, t) \right) \right\|_2^2
% \right]
% \end{equation}

% \begin{equation}
% \mathcal{L}_{\text{focus}} = 
% \mathbb{E}_{x_0, \epsilon, t, c_{\text{ID}}, c_{\text{LMK}}, c_{\text{EMO}}} \left[ 
% \left\| M \odot \left( \epsilon - \epsilon_\theta(x_t, t, c_{\text{ID}}, c_{\text{LMK}}, c_{\text{EMO}}) \right) \right\|_2^2
% \right]
% \end{equation}

% \begin{equation}
% \mathcal{L}_{\text{total}} = \mathcal{L}_{\text{diffusion}} + \lambda_{focus} \cdot \mathcal{L}_{\text{focus}} + \lambda_{AD} \cdot \mathcal{L}_{\text{AD}}
% \end{equation}

\begin{equation}
\mathcal{L}_{\text{total}} = \mathcal{L}_{\text{diffusion}} + \lambda_{\text{AD}} \cdot \mathcal{L}_{\text{AD}}
\end{equation}

% $\mathcal{L}_{\text{focus}}$ is the loss that focuses on the face region using the mask $M$. The coefficients $\lambda_{\text{focus}}$ and $\lambda_{\text{AD}}$ control the magnitude of each loss. 
\noindent
The coefficient $\lambda_{\text{AD}}$ control the magnitude of AD Loss.
In this stage, only the ID extractor, each control module, and the corresponding key and value layers of the cross-attention are trained, while the other components are frozen. For the ID extractor, only the CLIP encoder part is trained to extract ID features, whereas the ArcFace model for the ID token is kept fixed.

\subsection{Inference}
\label{sec: method_infer}
During inference, classifier-free guidance~\cite{ho2022classifier-free-guidance} is employed. Specifically, it is formulated as follows:

\begin{equation}
\begin{aligned}
\hat{\epsilon}_{\theta}(x_t, c_{\text{ID}}, c_{\text{LMK}}, c_{\text{EMO}}, t) &= \epsilon_{\theta}(x_t, t) + w_{\text{ID}} \cdot \left( \epsilon_{\theta}(x_t, c_{\text{ID}}, t) - \epsilon_{\theta}(x_t, t) \right) \\
&+ w_{\text{LMK}} \cdot \left( \epsilon_{\theta}(x_t, c_{\text{ID}}, c_{\text{LMK}}, t) - \epsilon_{\theta}(x_t, c_{\text{ID}}, t) \right) \\
&+ w_{\text{EMO}} \cdot \left( \epsilon_{\theta}(x_t, c_{\text{ID}}, c_{\text{EMO}}, t) - \epsilon_{\theta}(x_t, c_{\text{ID}}, t) \right)
\end{aligned}
\end{equation}

\noindent
Here, $ w_{\text{ID}}, w_{\text{LMK}}, w_{\text{EMO}} $ represent the respective control scales for each modality.
When $ c_{\text{LMK}} $ and $ c_{\text{EMO}} $ are dropped, the connections are completely cut off, effectively removing the features. 
As a result, we do not need to learn special embeddings to represent the absence of conditions.
% This eliminates the need to train the zero embeddings.

\section{Experiments}

\subsection{Experiments Setup}
\textbf{Dataset.}
% For training, we used approximately 70,000 images from the FFHQ~\cite{karras2019style_FFHQ} dataset, which provides diverse face images in terms of identity (ID) across various ethnicities, genders, and ages. However, it lacks sufficient diversity in terms of facial poses, expressions, and emotions, which are essential for control-based learning. To address this limitation, additional images were extracted from the FEED~\cite{drobyshev2024emoportraits} and MEAD~\cite{kaisiyuan2020mead} video datasets. Specifically, each video in the FEED dataset was converted into frames at 1-second intervals, followed by face detection and cropping to a standardized size. A similar process was applied to the MEAD dataset, though we randomly sampled approximately one-tenth of its videos due to its large scale. This combination resulted in a dataset of approximately 130,000 facial images.
% This augmented dataset not only increases the diversity in terms of identity but also improves the variation in facial poses, expressions, and emotions. 
For training, we use 70,000 images from FFHQ~\cite{karras2019style_FFHQ} for diverse identities. To enrich pose, expression, and emotion variation, we augment the dataset with frames extracted from FEED~\cite{drobyshev2024emoportraits} and a subset of MEAD~\cite{kaisiyuan2020mead} videos (sampled at 1-second intervals and cropped). This results in ~130,000 images with enhanced diversity in both identity and facial attributes.
% Corresponding facial landmarks for each image were generated using MediaPipe~\cite{lugaresi2019mediapipe}, and emotion vectors were extracted using Emonet~\cite{gerczuk2021emonet}, a pretrained emotion recognition model.

\noindent
\textbf{Implementation.} 
Our model is based on Arc2Face~\cite{papantoniou2024arc2face} and trained in two stages (100k/50k steps) using AdamW~\cite{loshchilov2017decoupled_AdamW}, batch size 16, and $\lambda_{\text{AD}}{=}0.0005$.
% We used Arc2Face~\cite{papantoniou2024arc2face}, a Stable Diffusion-based identity-conditional model, as the base model for our training.
% We trained our model in two stages (100k/50k steps) with AdamW~\cite{loshchilov2017decoupled_AdamW}, batch size 16, and $\lambda_{\text{AD}}{=}0.0005$. 
Inference uses DDPM~\cite{ddpm} with classifier-free guidance scales $w_{\text{ID}}{=}2.5$, $w_{\text{LMK}}{=}2.0$, and $w_{\text{EMO}}{=}2.0$. Details are in the supplementary material.

\subsection{Evaluation Metrics}
\textbf{Control Evaluation.} 
To evaluate control accuracy over pose, expression, and emotion, we construct a benchmark using CelebA~\cite{liu2015deep_celeb_dataset}. We select 100 identity images and 100 diverse target images in pose and expression, forming 2,000 identity-target pairs for generation and evaluation. We use the following metrics: (1) \textbf{Expression(Exp)/Pose RMSE}: RMSE of FLAME parameters~\cite{li2017learning_FLAME} extracted with EMOCA v2~\cite{danvevcek2022emoca, filntisis2022visual-emoca} to measure alignment with the target; (2) \textbf{ID Similarity}: cosine similarity from FaceNet~\cite{schroff2015facenet} to assess identity preservation; (3) \textbf{FID}: Fréchet Inception Distance~\cite{heusel2017gans_FID, seitzer2020pytorchfid} against 2,000 FFHQ images to evaluate realism; (4) \textbf{Valence/Arousal(V/A) RMSE}: RMSE of Valence/Arousal scores from Emonet to measure emotional consistency; (5) \textbf{Emotion Dist RMSE/Class ACC}: RMSE and classification accuracy of 8-class emotion distributions from Emonet~\cite{gerczuk2021emonet} to evaluate categorical emotion similarity. Baselines include Arc2Face+ControlNet~\cite{papantoniou2024arc2face} and CapHuman~\cite{liang2024caphuman}, which allow explicit pose and expression control.
\begin{table*}[h]
\centering
\resizebox{\textwidth}{!}{
\begin{tabular}{lcccccccccc}
\toprule
\textbf{Method} & 
\textbf{Exp-RMSE} ↓ & 
\textbf{Pose-RMSE} ↓ & 
\textbf{V-RMSE} ↓ & 
\textbf{A-RMSE} ↓ & 
\textbf{Dist-RMSE} ↓ & 
\textbf{Class-ACC(\%)} ↑ & 
\textbf{ID-Sim} ↑ & 
\textbf{FID} ↓ & 
\textbf{\#Extra-Params} ↓ \\
\midrule
Arc2Face+ControlNet & 8.12 & 0.130 & 0.360 & 0.247 & 0.86 & 35.1 & \textbf{0.76} & 44.64 & 361M \\
CapHuman & 7.82 & 0.166 & 0.428 & 0.224 & 0.87 & 34.6 & 0.71 & 88.80 & 361M \\
% Ours-Pre-trained(LMK) & 6.79 & 0.11 & 0.25 & 0.22 & 0.68 & 0.487 & 0.72 & 49.725 & 11.7M \\
% Ours-Pre-trained & 6.91 & 0.12 & 0.21 & 0.21 & 0.56 & 0.590 & 0.59 & 44.919 & 18M \\
% Ours(LMK) & 6.67 & 0.11 & 0.25 & 0.20 & 0.63 & 0.53 & 0.69 & 44.919 & 11.7M \\
% Ours & \textbf{6.67} & \textbf{0.11} & \textbf{0.21} & \textbf{0.17} & \textbf{0.48} & \textbf{0.632} & 0.67 & 44.727 & 18M \\
FaceCrafter-Pre-trained(LMK) & 7.11 & \underline{0.124} & 0.267 & 0.221 & 0.71 & 46.5 & \underline{0.75} & 46.85 & \textbf{11.7M} \\
FaceCrafter-Pre-trained(LMK+EMO) & 7.12 & 0.128 & \underline{0.218} & 0.205 & \underline{0.59} & \underline{56.5} & 0.72 & 53.25 & \underline{18M} \\
FaceCrafter(LMK) & \textbf{6.78} & \textbf{0.119} & 0.250 & \underline{0.203} & 0.63 & 52.4 & 0.71 & \textbf{38.80} & \textbf{11.7M} \\
FaceCrafter(LMK+EMO) & \underline{6.81} & 0.126 & \textbf{0.217} & \textbf{0.184} & \textbf{0.52} & \textbf{61.2} & 0.69 & \underline{41.54} & \underline{18M} \\
\bottomrule
\end{tabular}
}
\caption{Quantitative comparison of different methods in terms of controllability, identity preservation, and overall generation quality. Evaluations are performed on 2,000 image pairs, each consisting of an ID image and a target image, both sampled from CelebA~\cite{liu2015deep_celeb_dataset}. The rightmost column shows the number of additional parameters introduced for control by each method.
"Pre-trained" indicates the result after the Pre-trained stage. "LMK" uses only the LMK control module, and "LMK+EMO" uses both the LMK and EMO control modules.
Arrows indicate whether higher (↑) or lower (↓) values are preferred. The best results are shown in bold. The second-best results are underlined. }
\label{tab:control_eval_table}
\end{table*}
% \vspace{-70mm}
\begin{table}[H]
\centering
\resizebox{\textwidth}{!}{%
\begin{tabular}{lcccccccc}
\toprule
\multirow{2}{*}{Method} & \multicolumn{2}{c}{LPIPS $\uparrow$} & \multicolumn{2}{c}{Exp. ($\ell_2$) $\uparrow$} & \multicolumn{2}{c}{Pose ($\ell_2$) $\uparrow$} & \multicolumn{2}{c}{FID $\downarrow$} \\
\cmidrule(lr){2-3} \cmidrule(lr){4-5} \cmidrule(lr){6-7} \cmidrule(lr){8-9}
 & Synth & AgeDB & Synth & AgeDB & Synth & AgeDB & Synth & AgeDB \\
\midrule
FastComposer        & 0.389 & 0.487 & 3.597 & 4.678 & 0.163 & 0.225 & 13.517 & 31.736 \\
PhotoMaker          & 0.410 & 0.424 & 3.920 & 4.283 & 0.167 & 0.165 & 13.295 & 8.410 \\
InstantID           & 0.386 & 0.437 & 3.733 & 4.569 & 0.059 & 0.082 & 22.859 & 18.598 \\
IPA-FaceID (SDXL)   & 0.402 & 0.462 & 4.648 & 5.812 & 0.181 & 0.197 & 7.104  & 24.105 \\
IPA-FaceID-Plus     & 0.320 & 0.384 & 2.706 & 3.518 & 0.150 & 0.194 & 14.880 & 11.817 \\
IPA-FaceID-Plusv2   & 0.356 & 0.429 & 3.147 & 4.092 & 0.185 & 0.236 & 9.752  & 10.798 \\
Arc2Face            & 0.506 & 0.508 & 6.375 & 5.966 & 0.317 & 0.273 & 5.673 & 6.628 \\
% Ours                & \textbf{0.536} & \textbf{0.563} & \textbf{6.580} & \textbf{6.652} & \textbf{0.403} & \textbf{0.398} & \textbf{2.257} & \textbf{6.339} \\
FaceCrafter                & \textbf{0.552} & \textbf{0.589} & \textbf{6.971} & \textbf{6.930} & \textbf{0.429} & \textbf{0.424} & \textbf{2.112} & \textbf{6.501} \\
\bottomrule
\end{tabular}
}
\caption{Comparison of identity-conditional models on Synth-500 and AgeDB-500 datasets. Higher is better for LPIPS, Exp., and Pose; lower is better for FID.}
\label{tab:id_conditional_comparison}
\end{table}

\noindent
\textbf{Diversity Evaluation as ID-conditional foundation model.} For comparison as an ID-conditional generation model, we refer to the Arc2Face~\cite{papantoniou2024arc2face} evaluation framework. \textbf{It should be noted that, in this setting, Landmark/Emotion Control Modules are not used.} To ensure a fair comparison, we adopt the same evaluation datasets as in Arc2Face, namely Synth-500 (collected by~\cite{Synth-500}) and AgeDB-500~\cite{moschoglou2017agedb}. 
As baselines for comparison, we use Arc2Face as well as other state-of-the-art identity-conditional generation methods such as FastComposer~\cite{xiao2024fastcomposer}, PhotoMaker~\cite{li2024photomaker}, IP-Adapter-FaceID~\cite{ye2023ip-adapter, ye2024ipadapter_git}, and InstantID~\cite{wang2024instantid_big_model}.
% As baselines for comparison, we include Arc2Face and other state-of-the-art identity-conditional generation methods, including FastComposer~\cite{xiao2024fastcomposer}, PhotoMaker~\cite{li2024photomaker}, IP-Adapter-FaceID~\cite{ye2023ip-adapter, ye2024ipadapter_git}, and InstantID~\cite{wang2024instantid_big_model}.
To evaluate performance, we use the following metrics: LPIPS~\cite{zhang2018unreasonable_LPIPS} to measure diversity among generated samples, Expression/Pose diversity by computing the standard deviation of corresponding FLAME model parameters across samples, and FID to assess the realism of the generated images. 
% Note that during this evaluation, only the ID extractor is used as a control module, while other control adapters (e.g., for LMK or emotion) are not employed.

\noindent
\textbf{Evaluation via User Study.} To obtain accurate human-centric evaluations, we conduct a user study with 21 participants. They evaluate three aspects: diversity, controllability, and identity preservation. For diversity and controllability, evaluations are separately performed on two attributes — facial pose and expression (including emotion). For each evaluation, 30 sets of images are prepared, and each participant is randomly assigned 5 sets to assess. They are asked to select the most favorable image for each evaluation metric from a set of images generated by different methods, presented in random order. The results are reported as the selection rate for each method.

% \subsection{Experiments Results}
% \textbf{Quantitative Results:} Table~\ref{tab:control_eval_table} summarizes the control evaluation results. Table~\ref{tab:control_eval_table} summa-
\begin{table}[t]
\centering
\resizebox{\textwidth}{!}{%
\begin{tabular}{lcccccccc}
\toprule
\multirow{2}{*}{Method} & \multicolumn{2}{c}{LPIPS $\uparrow$} & \multicolumn{2}{c}{Exp. ($\ell_2$) $\uparrow$} & \multicolumn{2}{c}{Pose ($\ell_2$) $\uparrow$} & \multicolumn{2}{c}{FID $\downarrow$} \\
\cmidrule(lr){2-3} \cmidrule(lr){4-5} \cmidrule(lr){6-7} \cmidrule(lr){8-9}
 & Synth & AgeDB & Synth & AgeDB & Synth & AgeDB & Synth & AgeDB \\
\midrule
% Ours                & \textbf{0.536} & \textbf{0.563} & \textbf{6.580} & \textbf{6.652} & \textbf{0.403} & \textbf{0.398} & \textbf{2.257} & \textbf{6.339} \\
% Ours                & 0.536 & 0.563 & \textbf{6.580} & \textbf{6.652} & 0.403 & \textbf{0.398} & 2.257 & 6.339 \\
w/o AD Loss            & 0.538 & 0.564 & 6.540 & 6.430 & 0.411 & 0.397 & 2.140 & \textbf{5.381} \\
Ours                & \textbf{0.552} & \textbf{0.589} & \textbf{6.971} & \textbf{6.930} & \textbf{0.429} & \textbf{0.424} & \textbf{2.112} & 6.501 \\
\bottomrule
\end{tabular}
}
\caption{Ablation study on the effect of AD Loss.}
\label{tab:ablation_study}
\end{table}
\begin{table*}[t]
\centering
\begin{minipage}{0.45\linewidth}
\centering
\resizebox{\textwidth}{!}{%
\begin{tabular}{lccc}
\toprule
Method & Pose-Control(\%) $\uparrow$ & Exp-Control(\%) $\uparrow$ & ID-Sim(\%) $\uparrow$ \\
\midrule
Arc2Face+ControlNet & 15.2 & 21.0 & \textbf{36.9} \\
CapHuman & 6.7 & 5.7 & 32.4 \\
FaceCrafter & \textbf{78.1} & \textbf{73.3} & 30.6\\
\bottomrule
\end{tabular}
}
\caption{User study on pose/expression control and ID preservation.}
\label{tab:user_study_con}
\end{minipage}
\hfill
\begin{minipage}{0.45\linewidth}
\centering
\resizebox{\textwidth}{!}{%
\begin{tabular}{lcc}
\toprule
Method & Pose-Diversity(\%) $\uparrow$ & Exp-Diversity(\%) $\uparrow$ \\
\midrule
Arc2Face &  22.9 & 21.9 \\
FaceCrafter &  \textbf{77.1} & \textbf{78.1} \\
\bottomrule
\end{tabular}
}
\caption{User study on pose/expression diversity.}
\label{tab:user_study_div}
\end{minipage}
\end{table*}
\begin{figure}[H]
    \centering
    \includegraphics[width=0.60\linewidth]{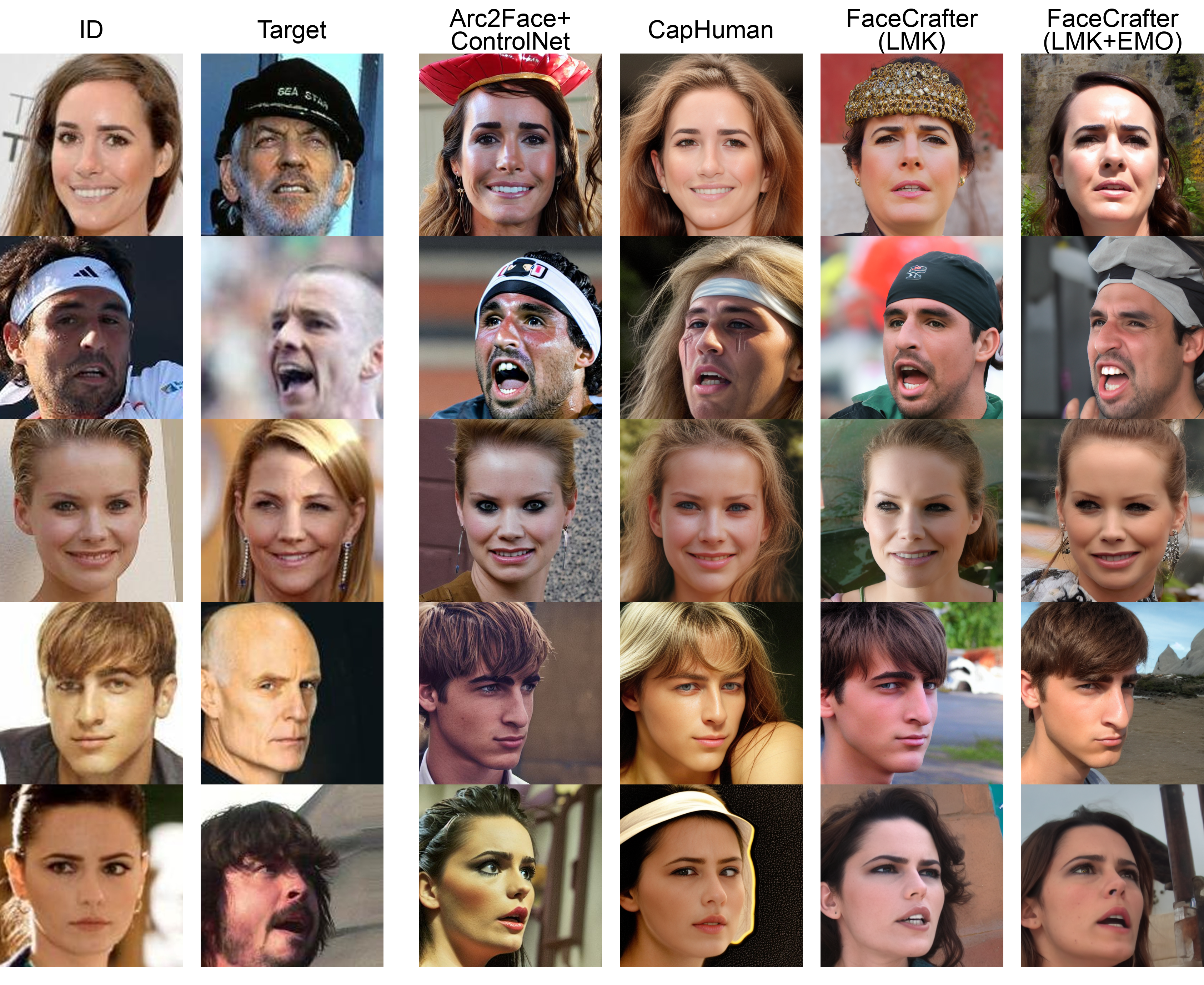}
    \caption{
    Qualitative comparison where the ID is guided to match the target's pose and expression. We compare our method with Arc2Face+ControlNet~\cite{papantoniou2024arc2face} and CapHuman~\cite{liang2024caphuman}.
    % Qualitative comparison of image generation where the ID image (first column) is controlled to match the pose and expression of the target image (second column). We compare our method with two baselines: Arc2Face+ControlNet~\cite{papantoniou2024arc2face} and CapHuman~\cite{liang2024caphuman}. 
    % Our methods, Ours (LMK only) and Ours, demonstrate superior control over both pose and expression, particularly achieving accurate emotion control.
    }
    \label{fig:generation control example on each methods}
\end{figure}

% con_pose results
% {'our': 0.5523809523809524, 'arc': 0.1523809523809524, 'our_lmk': 0.22857142857142856, 'cap': 0.06666666666666667, 'ours_all': 0.780952380952381}
% #############
% con_emo results
% {'our_lmk': 0.3238095238095238, 'arc': 0.20952380952380953, 'our': 0.4095238095238095, 'cap': 0.05714285714285714, 'ours_all': 0.7333333333333334}
% #############
% id results
% {'arc': 0.36936936936936937, 'our_lmk': 0.15315315315315314, 'our': 0.15315315315315314, 'cap': 0.32432432432432434, 'ours_all': 0.3063063063063063}
% \noindent
\subsection{Experiments Results}
\textbf{Quantitative Results.} Table~\ref{tab:control_eval_table} summarizes the control evaluation results. 
Our method outperforms existing methods in pose, expression, and emotion control, using only 5\% of the additional parameters. Notably, emotion class accuracy improves by 25\% over baselines, with a further 9\% gain from Emotion Control Module. Although ID similarity is slightly lower, it remains high despite variations such as eyeglasses and dataset limitations (see the supplementary material for further discussion). Our method also achieves the best FID score, indicating high image quality. Compared to the Pre-training Stage, Joint Disentanglement Stage improves overall control-performance.
% As shown in Table~\ref{tab:id_conditional_comparison}, our method yields higher diversity across all metrics. Table~\ref{tab:ablation_study} shows that the AD loss notably enhances expression diversity, though diversity remains higher than baselines even without it. 
% Under identity-only conditioning, our method still demonstrates strong diversity across all metrics, as shown in Table~\ref{tab:id_conditional_comparison}. This indicates the effectiveness of our approach in generating varied outputs even without explicit other control signals. 
% Next, we evaluate our model under identity-only conditioning to assess its capability as an identity-conditional foundation model. As shown in Table~\ref{tab:id_conditional_comparison}, our method significantly outperforms existing approaches across multiple diversity metrics, including LPIPS, Expression, and Pose variance. In particular, we observe substantial improvements in the diversity of facial expressions and head poses, indicating that our model can naturally produce varied and realistic outputs even without additional control signals. Moreover, our method also achieves superior FID scores compared to baselines, suggesting that the generated images are not only more diverse but also of higher perceptual quality. These results demonstrate the strong generative capacity and robustness of our approach.
Next, we evaluate our model under identity-only conditioning to assess its capability as an identity-conditional foundation model. As shown in Table~\ref{tab:id_conditional_comparison}, our method achieves superior diversity in LPIPS, Expression, and Pose, while also outperforming baselines in FID. These results highlight the model’s strong generative ability and robustness, even without explicit control signals.

\noindent
\textbf{Ablation Study.}
% Table~\ref{tab:ablation_study} further shows that the AD loss notably enhances expression diversity, though diversity remains higher than baselines even without it.
Table~\ref{tab:ablation_study} presents an ablation study examining the impact of the AD Loss. The results demonstrate that introducing the AD Loss substantially boosts the overall diversity of generated images, with particularly pronounced improvements in expression diversity. This indicates that AD Loss effectively facilitates the disentanglement of identity and expression features, allowing for more varied and expressive outputs. Interestingly, even without AD Loss, our model still achieves higher diversity compared to baseline methods, showing the strength of our underlying framework. The FID score on AgeDB is slightly better without AD Loss, indicating that although it boosts diversity, it may cause some training instability and slight drops in image quality.

\noindent
\textbf{User Study.} 
Based on the user study results shown in Tables~\ref{tab:user_study_con} and~\ref{tab:user_study_div}, our method demonstrates a significant advantage in terms of controllability and diversity. For ID preservation, although no method shows clear superiority, our approach achieves comparable performance, indicating that it maintains identity consistency while enabling flexible generation control. Importantly, the user study complements score-based metrics, providing human-centered evidence that supports the reliability of our method.

\begin{figure}[t]
    \centering
    \includegraphics[width=0.90\linewidth]{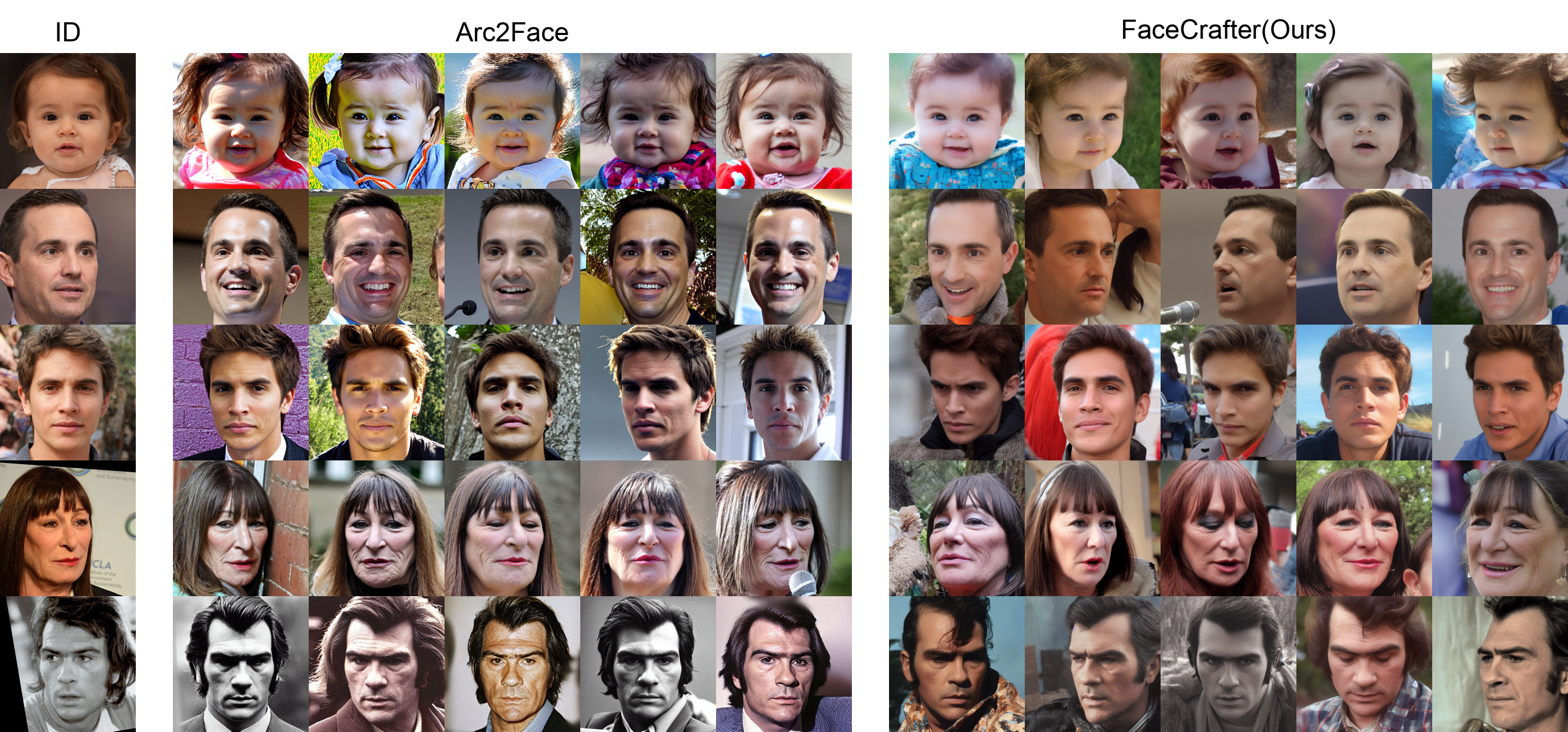}
    \caption{Examples of image generation conditioned only on ID condition. 
    % Left: ID images; Middle: Arc2Face; Right: FaceCrafter(Ours).
    }
    \label{fig:diversity example}
\end{figure}

\noindent
\textbf{Qualitative Results.}
% For qualitative evaluation, Figure~\ref{fig:generation control example on each methods} shows example images generated by each method when controlling the target face orientation and expression based on various IDs. Our method achieves more accurate control to match the target compared to other methods. Moreover, when comparing the results of LMK only and LMK + EMO, the latter provides better control over emotions.
% Figure~\ref{fig:diversity example} presents examples generated under the ID-only condition to assess diversity. For each ID, five images are generated. Overall, our method exhibits greater diversity in pose, expression, and emotion compared to the baseline, Arc2Face. For instance, in row 3, all results from Arc2Face show similar expressions to the ID, while our method generates a wider variety of poses, expressions, and emotions. Additional qualitative results are provided in the supplementary material.
For qualitative evaluation, Figure~\ref{fig:generation control example on each methods} shows that our method more accurately controls pose and expression compared to other methods. Moreover, when comparing the results of LMK only and LMK + EMO, the latter provides better control over emotions.
Figure~\ref{fig:diversity example} shows samples under ID-only conditioning. Our method generates more diverse poses, expressions, and emotions than Arc2Face. For example, in row 3, Arc2Face shows similar expressions across samples, while ours captures greater variation. More results are in the supplementary material.
\section{Conclusions}
% In this work, we proposed a controllable ID-conditional diffusion framework that introduces lightweight control modules into the cross-attention layers of the base model, enabling controllable manipulation of facial pose, expression, and emotion while preserving identity. We further presented a novel training strategy that jointly optimizes the control modules and the identity extractor to encourage disentanglement between identity features and non-identity control features, resulting in improved controllability and generative diversity. Through comprehensive quantitative, qualitative, and user study evaluations, We demonstrated that our method enables fine-grained attribute control, while also achieving high diversity even under identity-only conditioning. We believe that our approach offers a promising foundation for future research in both generative and recognition-based face-related tasks.
% We proposed a controllable ID-conditional diffusion framework that introduces lightweight control modules into cross-attention layers, enabling manipulation of facial pose, expression, and emotion while preserving identity. A novel training strategy jointly optimizes the control modules and identity extractor to disentangle identity and non-identity features, improving both controllability and diversity. Comprehensive evaluations demonstrate fine-grained control and high diversity, providing a strong foundation for future face generation and recognition research.
We propose a controllable ID-conditional diffusion framework with lightweight control modules in cross-attention layers, enabling manipulation of facial pose, expression, and emotion while preserving identity. Our novel training strategy disentangles identity and non-identity features, improving controllability and diversity. Evaluations show fine-grained control and high diversity, supporting future face generation research.
% \noindent
% \textbf{Acknowledgments.}
% This work was supported by JST SPRING, Japan Grant Number JPMJSP2180.
\section*{Acknowledgments}
This work was supported by JST SPRING, Japan Grant Number JPMJSP2180.

\bibliography{egbib}

\begin{thebibliography}{67}
\providecommand{\natexlab}[1]{#1}
\providecommand{\url}[1]{\texttt{#1}}
\expandafter\ifx\csname urlstyle\endcsname\relax
  \providecommand{\doi}[1]{doi: #1}\else
  \providecommand{\doi}{doi: \begingroup \urlstyle{rm}\Url}\fi

\bibitem[Azari and Lim(2024)]{azari2024emostyle}
Bita Azari and Angelica Lim.
\newblock Emostyle: One-shot facial expression editing using continuous emotion
  parameters.
\newblock In \emph{Proceedings of the IEEE/CVF Winter Conference on
  Applications of Computer Vision}, pages 6385--6394, 2024.

\bibitem[Bansal et~al.(2023)Bansal, Chu, Schwarzschild, Sengupta, Goldblum,
  Geiping, and Goldstein]{bansal2023universal_diffusion}
Arpit Bansal, Hong-Min Chu, Avi Schwarzschild, Soumyadip Sengupta, Micah
  Goldblum, Jonas Geiping, and Tom Goldstein.
\newblock Universal guidance for diffusion models.
\newblock In \emph{Proceedings of the IEEE/CVF Conference on Computer Vision
  and Pattern Recognition}, pages 843--852, 2023.

\bibitem[Chen et~al.(2023)Chen, Zhao, Liu, Ding, Song, Wang, Wang, Yang, Liu,
  Du, et~al.]{chen2023photoverse_similar_ip-adapter}
Li~Chen, Mengyi Zhao, Yiheng Liu, Mingxu Ding, Yangyang Song, Shizun Wang,
  Xu~Wang, Hao Yang, Jing Liu, Kang Du, et~al.
\newblock Photoverse: Tuning-free image customization with text-to-image
  diffusion models.
\newblock \emph{arXiv preprint arXiv:2309.05793}, 2023.

\bibitem[Chen et~al.(2024)Chen, Fang, Liu, He, Huang, and
  Mao]{chen2024dreamidentity}
Zhuowei Chen, Shancheng Fang, Wei Liu, Qian He, Mengqi Huang, and Zhendong Mao.
\newblock Dreamidentity: enhanced editability for efficient face-identity
  preserved image generation.
\newblock In \emph{Proceedings of the AAAI Conference on Artificial
  Intelligence}, volume~38, pages 1281--1289, 2024.
\newblock 2.

\bibitem[Choi et~al.(2018)Choi, Choi, Kim, Ha, Kim, and
  Choo]{choi2018stargan_expedit}
Yunjey Choi, Minje Choi, Munyoung Kim, Jung-Woo Ha, Sunghun Kim, and Jaegul
  Choo.
\newblock Stargan: Unified generative adversarial networks for multi-domain
  image-to-image translation.
\newblock In \emph{Proceedings of the IEEE conference on computer vision and
  pattern recognition}, pages 8789--8797, 2018.

\bibitem[Dan{\v{e}}{\v{c}}ek et~al.(2022)Dan{\v{e}}{\v{c}}ek, Black, and
  Bolkart]{danvevcek2022emoca}
Radek Dan{\v{e}}{\v{c}}ek, Michael~J Black, and Timo Bolkart.
\newblock Emoca: Emotion driven monocular face capture and animation.
\newblock In \emph{Proceedings of the IEEE/CVF Conference on Computer Vision
  and Pattern Recognition}, pages 20311--20322, 2022.

\bibitem[Deng et~al.(2019)Deng, Guo, Xue, and Zafeiriou]{deng2019arcface}
Jiankang Deng, Jia Guo, Niannan Xue, and Stefanos Zafeiriou.
\newblock Arcface: Additive angular margin loss for deep face recognition.
\newblock In \emph{Proceedings of the IEEE/CVF conference on computer vision
  and pattern recognition}, pages 4690--4699, 2019.

\bibitem[Ding et~al.(2018)Ding, Sricharan, and
  Chellappa]{ding2018exprgan_ExprGan_expedit}
Hui Ding, Kumar Sricharan, and Rama Chellappa.
\newblock Exprgan: Facial expression editing with controllable expression
  intensity.
\newblock In \emph{Proceedings of the AAAI conference on artificial
  intelligence}, volume~32, 2018.
\newblock 1.

\bibitem[Drobyshev et~al.(2024)Drobyshev, Casademunt, Vougioukas, Landgraf,
  Petridis, and Pantic]{drobyshev2024emoportraits}
Nikita Drobyshev, Antoni~Bigata Casademunt, Konstantinos Vougioukas, Zoe
  Landgraf, Stavros Petridis, and Maja Pantic.
\newblock Emoportraits: Emotion-enhanced multimodal one-shot head avatars.
\newblock In \emph{Proceedings of the IEEE/CVF Conference on Computer Vision
  and Pattern Recognition}, pages 8498--8507, 2024.

\bibitem[Filntisis et~al.(2022)Filntisis, Retsinas, Paraperas-Papantoniou,
  Katsamanis, Roussos, and Maragos]{filntisis2022visual-emoca}
Panagiotis~P Filntisis, George Retsinas, Foivos Paraperas-Papantoniou,
  Athanasios Katsamanis, Anastasios Roussos, and Petros Maragos.
\newblock Visual speech-aware perceptual 3d facial expression reconstruction
  from videos.
\newblock \emph{arXiv preprint arXiv:2207.11094}, 2022.

\bibitem[Gal et~al.(2023{\natexlab{a}})Gal, Alaluf, Atzmon, Patashnik, Bermano,
  Chechik, and Cohen-or]{gal2022image_textual_inversion}
Rinon Gal, Yuval Alaluf, Yuval Atzmon, Or~Patashnik, Amit~Haim Bermano, Gal
  Chechik, and Daniel Cohen-or.
\newblock An image is worth one word: Personalizing text-to-image generation
  using textual inversion.
\newblock In \emph{The Eleventh International Conference on Learning
  Representations}, 2023{\natexlab{a}}.
\newblock URL \url{https://openreview.net/forum?id=NAQvF08TcyG}.

\bibitem[Gal et~al.(2023{\natexlab{b}})Gal, Arar, Atzmon, Bermano, Chechik, and
  Cohen-Or]{gal2023encoder_personalized}
Rinon Gal, Moab Arar, Yuval Atzmon, Amit~H Bermano, Gal Chechik, and Daniel
  Cohen-Or.
\newblock Encoder-based domain tuning for fast personalization of text-to-image
  models.
\newblock \emph{ACM Transactions on Graphics (TOG)}, 42\penalty0 (4):\penalty0
  1--13, 2023{\natexlab{b}}.

\bibitem[Gerczuk et~al.(2021)Gerczuk, Amiriparian, Ottl, and
  Schuller]{gerczuk2021emonet}
Maurice Gerczuk, Shahin Amiriparian, Sandra Ottl, and Bj{\"o}rn~W Schuller.
\newblock Emonet: A transfer learning framework for multi-corpus speech emotion
  recognition.
\newblock \emph{IEEE Transactions on Affective Computing}, 14\penalty0
  (2):\penalty0 1472--1487, 2021.

\bibitem[Goodfellow et~al.(2014)Goodfellow, Pouget-Abadie, Mirza, Xu,
  Warde-Farley, Ozair, Courville, and Bengio]{goodfellow2014generative_GAN}
Ian~J Goodfellow, Jean Pouget-Abadie, Mehdi Mirza, Bing Xu, David Warde-Farley,
  Sherjil Ozair, Aaron Courville, and Yoshua Bengio.
\newblock Generative adversarial nets.
\newblock \emph{Advances in neural information processing systems}, 27, 2014.

\bibitem[H{\"a}rk{\"o}nen et~al.(2020)H{\"a}rk{\"o}nen, Hertzmann, Lehtinen,
  and Paris]{harkonen2020ganspace}
Erik H{\"a}rk{\"o}nen, Aaron Hertzmann, Jaakko Lehtinen, and Sylvain Paris.
\newblock Ganspace: Discovering interpretable gan controls.
\newblock \emph{Advances in neural information processing systems},
  33:\penalty0 9841--9850, 2020.

\bibitem[He et~al.(2024)He, Sun, Juefei-Xu, Ma, Ramchandani, Cheung, Shah,
  Kalia, Subramanyam, Zareian, Chen, Jain, Zhang, Zhang, Sumbaly, Vajda, and
  Sinha]{he2024imagine_yourself_meta_big_model}
Zecheng He, Bo~Sun, Felix Juefei-Xu, Haoyu Ma, Ankit Ramchandani, Vincent
  Cheung, Siddharth Shah, Anmol Kalia, Harihar Subramanyam, Alireza Zareian,
  Li~Chen, Ankit Jain, Ning Zhang, Peizhao Zhang, Roshan Sumbaly, Peter Vajda,
  and Animesh Sinha.
\newblock Imagine yourself: Tuning-free personalized image generation.
\newblock \emph{CoRR}, abs/2409.13346, 2024.
\newblock URL \url{https://doi.org/10.48550/arXiv.2409.13346}.

\bibitem[Heusel et~al.(2017)Heusel, Ramsauer, Unterthiner, Nessler, and
  Hochreiter]{heusel2017gans_FID}
Martin Heusel, Hubert Ramsauer, Thomas Unterthiner, Bernhard Nessler, and Sepp
  Hochreiter.
\newblock Gans trained by a two time-scale update rule converge to a local nash
  equilibrium.
\newblock \emph{Advances in neural information processing systems}, 30, 2017.

\bibitem[Ho and Salimans(2021)]{ho2022classifier-free-guidance}
Jonathan Ho and Tim Salimans.
\newblock Classifier-free diffusion guidance.
\newblock In \emph{NeurIPS 2021 Workshop on Deep Generative Models and
  Downstream Applications}, 2021.
\newblock URL \url{https://openreview.net/forum?id=qw8AKxfYbI}.

\bibitem[Ho et~al.(2020)Ho, Jain, and Abbeel]{ddpm}
Jonathan Ho, Ajay Jain, and Pieter Abbeel.
\newblock Denoising diffusion probabilistic models.
\newblock \emph{Advances in neural information processing systems},
  33:\penalty0 6840--6851, 2020.

\bibitem[Hu et~al.(2018)Hu, Wu, Yu, He, and Sun]{hu2018pose_gan}
Yibo Hu, Xiang Wu, Bing Yu, Ran He, and Zhenan Sun.
\newblock Pose-guided photorealistic face rotation.
\newblock In \emph{Proceedings of the IEEE conference on computer vision and
  pattern recognition}, pages 8398--8406, 2018.

\bibitem[Huang et~al.(2008)Huang, Mattar, Berg, and
  Learned-Miller]{huang2008labeled_lfw_dataset}
Gary~B Huang, Marwan Mattar, Tamara Berg, and Eric Learned-Miller.
\newblock Labeled faces in the wild: A database forstudying face recognition in
  unconstrained environments.
\newblock In \emph{Workshop on faces in'Real-Life'Images: detection, alignment,
  and recognition}, 2008.

\bibitem[Huang et~al.(2017)Huang, Zhang, Li, and He]{huang2017beyond_pose_gan}
Rui Huang, Shu Zhang, Tianyu Li, and Ran He.
\newblock Beyond face rotation: Global and local perception gan for
  photorealistic and identity preserving frontal view synthesis.
\newblock In \emph{Proceedings of the IEEE international conference on computer
  vision}, pages 2439--2448, 2017.

\bibitem[Karras et~al.(2018)Karras, Aila, Laine, and
  Lehtinen]{karras2017progressive_celeb_hq}
Tero Karras, Timo Aila, Samuli Laine, and Jaakko Lehtinen.
\newblock Progressive growing of {GAN}s for improved quality, stability, and
  variation.
\newblock In \emph{International Conference on Learning Representations}, 2018.
\newblock URL \url{https://openreview.net/forum?id=Hk99zCeAb}.

\bibitem[Karras et~al.(2019{\natexlab{a}})Karras, Laine, and
  Aila]{karras2019style_FFHQ}
Tero Karras, Samuli Laine, and Timo Aila.
\newblock A style-based generator architecture for generative adversarial
  networks.
\newblock In \emph{Proceedings of the IEEE/CVF conference on computer vision
  and pattern recognition}, pages 4401--4410, 2019{\natexlab{a}}.

\bibitem[Karras et~al.(2019{\natexlab{b}})Karras, Laine, and
  Aila]{karras2019style_stylegan}
Tero Karras, Samuli Laine, and Timo Aila.
\newblock A style-based generator architecture for generative adversarial
  networks.
\newblock In \emph{Proceedings of the IEEE/CVF conference on computer vision
  and pattern recognition}, pages 4401--4410, 2019{\natexlab{b}}.

\bibitem[Karras et~al.(2020)Karras, Laine, Aittala, Hellsten, Lehtinen, and
  Aila]{karras2020analyzing_stylegan2}
Tero Karras, Samuli Laine, Miika Aittala, Janne Hellsten, Jaakko Lehtinen, and
  Timo Aila.
\newblock Analyzing and improving the image quality of stylegan.
\newblock In \emph{Proceedings of the IEEE/CVF conference on computer vision
  and pattern recognition}, pages 8110--8119, 2020.

\bibitem[Karras et~al.(2021)Karras, Aittala, Laine, H{\"a}rk{\"o}nen, Hellsten,
  Lehtinen, and Aila]{karras2021alias_stylegan3}
Tero Karras, Miika Aittala, Samuli Laine, Erik H{\"a}rk{\"o}nen, Janne
  Hellsten, Jaakko Lehtinen, and Timo Aila.
\newblock Alias-free generative adversarial networks.
\newblock \emph{Advances in neural information processing systems},
  34:\penalty0 852--863, 2021.

\bibitem[Li et~al.(2017{\natexlab{a}})Li, Deng, and Du]{li2017reliable_raf-db}
Shan Li, Weihong Deng, and JunPing Du.
\newblock Reliable crowdsourcing and deep locality-preserving learning for
  expression recognition in the wild.
\newblock In \emph{Proceedings of the IEEE conference on computer vision and
  pattern recognition}, pages 2852--2861, 2017{\natexlab{a}}.

\bibitem[Li et~al.(2017{\natexlab{b}})Li, Bolkart, Black, Li, and
  Romero]{li2017learning_FLAME}
Tianye Li, Timo Bolkart, Michael~J Black, Hao Li, and Javier Romero.
\newblock Learning a model of facial shape and expression from 4d scans.
\newblock \emph{ACM Trans. Graph.}, 36\penalty0 (6):\penalty0 194--1,
  2017{\natexlab{b}}.

\bibitem[Li et~al.(2024)Li, Cao, Wang, Qi, Cheng, and Shan]{li2024photomaker}
Zhen Li, Mingdeng Cao, Xintao Wang, Zhongang Qi, Ming-Ming Cheng, and Ying
  Shan.
\newblock Photomaker: Customizing realistic human photos via stacked id
  embedding.
\newblock In \emph{Proceedings of the IEEE/CVF conference on computer vision
  and pattern recognition}, pages 8640--8650, 2024.

\bibitem[Liang et~al.(2024)Liang, Ma, Zhu, Deng, and Yang]{liang2024caphuman}
Chao Liang, Fan Ma, Linchao Zhu, Yingying Deng, and Yi~Yang.
\newblock Caphuman: Capture your moments in parallel universes.
\newblock In \emph{Proceedings of the IEEE/CVF Conference on Computer Vision
  and Pattern Recognition}, pages 6400--6409, 2024.

\bibitem[Lindt et~al.(2019)Lindt, Barros, Siqueira, and
  Wermter]{lindt2019facial_expedit}
Alexandra Lindt, Pablo Barros, Henrique Siqueira, and Stefan Wermter.
\newblock Facial expression editing with continuous emotion labels.
\newblock In \emph{2019 14th IEEE International Conference on Automatic Face \&
  Gesture Recognition (FG 2019)}, pages 1--8. IEEE, 2019.

\bibitem[Liu et~al.(2024)Liu, Li, Sun, Ming, Fang, Wang, Zeng, and
  Liu]{liu2024ada_similar-ip-adapter}
Jia Liu, Changlin Li, Qirui Sun, Jiahui Ming, Chen Fang, Jue Wang, Bing Zeng,
  and Shuaicheng Liu.
\newblock Ada-adapter: Fast few-shot style personlization of diffusion model
  with pre-trained image encoder.
\newblock \emph{arXiv preprint arXiv:2407.05552}, 2024.

\bibitem[Liu et~al.(2015)Liu, Luo, Wang, and Tang]{liu2015deep_celeb_dataset}
Ziwei Liu, Ping Luo, Xiaogang Wang, and Xiaoou Tang.
\newblock Deep learning face attributes in the wild.
\newblock In \emph{Proceedings of the IEEE international conference on computer
  vision}, pages 3730--3738, 2015.

\bibitem[Loshchilov and Hutter(2019)]{loshchilov2017decoupled_AdamW}
Ilya Loshchilov and Frank Hutter.
\newblock Decoupled weight decay regularization.
\newblock In \emph{International Conference on Learning Representations}, 2019.
\newblock URL \url{https://openreview.net/forum?id=Bkg6RiCqY7}.

\bibitem[Lugaresi et~al.(2019)Lugaresi, Tang, Nash, McClanahan, Uboweja, Hays,
  Zhang, Chang, Yong, Lee, Chang, Hua, Georg, and
  Grundmann]{lugaresi2019mediapipe}
Camillo Lugaresi, Jiuqiang Tang, Hadon Nash, Chris McClanahan, Esha Uboweja,
  Michael Hays, Fan Zhang, Chuo{-}Ling Chang, Ming~Guang Yong, Juhyun Lee,
  Wan{-}Teh Chang, Wei Hua, Manfred Georg, and Matthias Grundmann.
\newblock Mediapipe: {A} framework for building perception pipelines.
\newblock \emph{CoRR}, abs/1906.08172, 2019.
\newblock URL \url{http://arxiv.org/abs/1906.08172}.

\bibitem[Luo et~al.(2020)Luo, Xue, and Feng]{luo2020ehanet}
Ling Luo, Dingyu Xue, and Xinglong Feng.
\newblock Ehanet: An effective hierarchical aggregation network for face
  parsing.
\newblock \emph{Applied Sciences}, 10\penalty0 (9):\penalty0 3135, 2020.

\bibitem[Mollahosseini et~al.(2017)Mollahosseini, Hasani, and
  Mahoor]{mollahosseini2017affectnet}
Ali Mollahosseini, Behzad Hasani, and Mohammad~H Mahoor.
\newblock Affectnet: A database for facial expression, valence, and arousal
  computing in the wild.
\newblock \emph{IEEE Transactions on Affective Computing}, 10\penalty0
  (1):\penalty0 18--31, 2017.

\bibitem[Moschoglou et~al.(2017)Moschoglou, Papaioannou, Sagonas, Deng, Kotsia,
  and Zafeiriou]{moschoglou2017agedb}
Stylianos Moschoglou, Athanasios Papaioannou, Christos Sagonas, Jiankang Deng,
  Irene Kotsia, and Stefanos Zafeiriou.
\newblock Agedb: the first manually collected, in-the-wild age database.
\newblock In \emph{proceedings of the IEEE conference on computer vision and
  pattern recognition workshops}, pages 51--59, 2017.

\bibitem[Nitzan et~al.(2022)Nitzan, Aberman, He, Liba, Yarom, Gandelsman,
  Mosseri, Pritch, and
  Cohen-Or]{nitzan2022mystyle_personalized_generative_prior}
Yotam Nitzan, Kfir Aberman, Qiurui He, Orly Liba, Michal Yarom, Yossi
  Gandelsman, Inbar Mosseri, Yael Pritch, and Daniel Cohen-Or.
\newblock Mystyle: A personalized generative prior.
\newblock \emph{ACM Transactions on Graphics (TOG)}, 41\penalty0 (6):\penalty0
  1--10, 2022.

\bibitem[Papantoniou et~al.(2024)Papantoniou, Lattas, Moschoglou, Deng, Kainz,
  and Zafeiriou]{papantoniou2024arc2face}
Foivos~Paraperas Papantoniou, Alexandros Lattas, Stylianos Moschoglou, Jiankang
  Deng, Bernhard Kainz, and Stefanos Zafeiriou.
\newblock Arc2face: A foundation model for id-consistent human faces.
\newblock In \emph{European Conference on Computer Vision}, pages 241--261.
  Springer, 2024.

\bibitem[Patashnik et~al.(2021)Patashnik, Wu, Shechtman, Cohen-Or, and
  Lischinski]{patashnik2021styleclip}
Or~Patashnik, Zongze Wu, Eli Shechtman, Daniel Cohen-Or, and Dani Lischinski.
\newblock Styleclip: Text-driven manipulation of stylegan imagery.
\newblock In \emph{Proceedings of the IEEE/CVF international conference on
  computer vision}, pages 2085--2094, 2021.

\bibitem[Peng et~al.(2024)Peng, Zhu, Jiang, Tai, Luo, Zhang, Lin, Jin, Wang,
  and Ji]{peng2024portraitbooth}
Xu~Peng, Junwei Zhu, Boyuan Jiang, Ying Tai, Donghao Luo, Jiangning Zhang, Wei
  Lin, Taisong Jin, Chengjie Wang, and Rongrong Ji.
\newblock Portraitbooth: A versatile portrait model for fast identity-preserved
  personalization.
\newblock In \emph{Proceedings of the IEEE/CVF Conference on Computer Vision
  and Pattern Recognition}, pages 27080--27090, 2024.

\bibitem[Pumarola et~al.(2018)Pumarola, Agudo, Martinez, Sanfeliu, and
  Moreno-Noguer]{pumarola2018ganimation_expedit}
Albert Pumarola, Antonio Agudo, Aleix~M Martinez, Alberto Sanfeliu, and
  Francesc Moreno-Noguer.
\newblock Ganimation: Anatomically-aware facial animation from a single image.
\newblock In \emph{Proceedings of the European conference on computer vision
  (ECCV)}, pages 818--833, 2018.

\bibitem[Radford et~al.(2021)Radford, Kim, Hallacy, Ramesh, Goh, Agarwal,
  Sastry, Askell, Mishkin, Clark, et~al.]{radford2021learning_CLIP}
Alec Radford, Jong~Wook Kim, Chris Hallacy, Aditya Ramesh, Gabriel Goh,
  Sandhini Agarwal, Girish Sastry, Amanda Askell, Pamela Mishkin, Jack Clark,
  et~al.
\newblock Learning transferable visual models from natural language
  supervision.
\newblock In \emph{International conference on machine learning}, pages
  8748--8763. PmLR, 2021.

\bibitem[Rombach et~al.(2022)Rombach, Blattmann, Lorenz, Esser, and
  Ommer]{rombach2022high_stable_diffusion}
Robin Rombach, Andreas Blattmann, Dominik Lorenz, Patrick Esser, and Bj{\"o}rn
  Ommer.
\newblock High-resolution image synthesis with latent diffusion models.
\newblock In \emph{Proceedings of the IEEE/CVF conference on computer vision
  and pattern recognition}, pages 10684--10695, 2022.

\bibitem[Ruiz et~al.(2023)Ruiz, Li, Jampani, Pritch, Rubinstein, and
  Aberman]{ruiz2023dreambooth}
Nataniel Ruiz, Yuanzhen Li, Varun Jampani, Yael Pritch, Michael Rubinstein, and
  Kfir Aberman.
\newblock Dreambooth: Fine tuning text-to-image diffusion models for
  subject-driven generation.
\newblock In \emph{Proceedings of the IEEE/CVF conference on computer vision
  and pattern recognition}, pages 22500--22510, 2023.

\bibitem[Ruiz et~al.(2024)Ruiz, Li, Jampani, Wei, Hou, Pritch, Wadhwa,
  Rubinstein, and Aberman]{ruiz2024hyperdreambooth}
Nataniel Ruiz, Yuanzhen Li, Varun Jampani, Wei Wei, Tingbo Hou, Yael Pritch,
  Neal Wadhwa, Michael Rubinstein, and Kfir Aberman.
\newblock Hyperdreambooth: Hypernetworks for fast personalization of
  text-to-image models.
\newblock In \emph{Proceedings of the IEEE/CVF conference on computer vision
  and pattern recognition}, pages 6527--6536, 2024.

\bibitem[Russell(1980)]{russell1980circumplex_Valence_Arousal}
James~A Russell.
\newblock A circumplex model of affect.
\newblock \emph{Journal of personality and social psychology}, 39\penalty0
  (6):\penalty0 1161, 1980.

\bibitem[Schroff et~al.(2015)Schroff, Kalenichenko, and
  Philbin]{schroff2015facenet}
Florian Schroff, Dmitry Kalenichenko, and James Philbin.
\newblock Facenet: A unified embedding for face recognition and clustering.
\newblock In \emph{Proceedings of the IEEE conference on computer vision and
  pattern recognition}, pages 815--823, 2015.

\bibitem[Seitzer(2020)]{seitzer2020pytorchfid}
M.~Seitzer.
\newblock pytorch-fid: Fid score for pytorch.
\newblock \url{https://github.com/mseitzer/pytorch-fid}, 2020.
\newblock Version 0.3.0, August 2020.

\bibitem[Shen et~al.(2020)Shen, Gu, Tang, and
  Zhou]{shen2020interpreting_InterFaceGAN}
Yujun Shen, Jinjin Gu, Xiaoou Tang, and Bolei Zhou.
\newblock Interpreting the latent space of gans for semantic face editing.
\newblock In \emph{Proceedings of the IEEE/CVF conference on computer vision
  and pattern recognition}, pages 9243--9252, 2020.

\bibitem[Synth-500()]{Synth-500}
Synth-500.
\newblock This person does not exist.
\newblock \url{https://thispersondoesnotexist.com/}, 2019.
\newblock Accessed: 2025-05-01.

\bibitem[Tang and Sebe(2022)]{tang2022facial_emo_gan}
Hao Tang and Nicu Sebe.
\newblock Facial expression translation using landmark guided gans.
\newblock \emph{IEEE Transactions on Affective Computing}, 13\penalty0
  (4):\penalty0 1986--1997, 2022.

\bibitem[Valevski et~al.(2023)Valevski, Lumen, Matias, and
  Leviathan]{valevski2023face0}
Dani Valevski, Danny Lumen, Yossi Matias, and Yaniv Leviathan.
\newblock Face0: Instantaneously conditioning a text-to-image model on a face.
\newblock In \emph{SIGGRAPH Asia 2023 Conference Papers}, pages 1--10, 2023.

\bibitem[Wang et~al.(2020)Wang, Wu, Song, Yang, Wu, Qian, He, Qiao, and
  Loy]{kaisiyuan2020mead}
Kaisiyuan Wang, Qianyi Wu, Linsen Song, Zhuoqian Yang, Wayne Wu, Chen Qian, Ran
  He, Yu~Qiao, and Chen~Change Loy.
\newblock Mead: A large-scale audio-visual dataset for emotional talking-face
  generation.
\newblock In \emph{ECCV}, August 2020.

\bibitem[Wang et~al.(2024)Wang, Bai, Wang, Qin, Chen, Li, Tang, and
  Hu]{wang2024instantid_big_model}
Qixun Wang, Xu~Bai, Haofan Wang, Zekui Qin, Anthony Chen, Huaxia Li, Xu~Tang,
  and Yao Hu.
\newblock Instantid: Zero-shot identity-preserving generation in seconds.
\newblock \emph{arXiv preprint arXiv:2401.07519}, 2024.

\bibitem[Xia et~al.(2021)Xia, Yang, Xue, and
  Wu]{xia2021tedigan_gan_face_control}
Weihao Xia, Yujiu Yang, Jing-Hao Xue, and Baoyuan Wu.
\newblock Tedigan: Text-guided diverse face image generation and manipulation.
\newblock In \emph{Proceedings of the IEEE/CVF conference on computer vision
  and pattern recognition}, pages 2256--2265, 2021.

\bibitem[Xiao et~al.(2024)Xiao, Yin, Freeman, Durand, and
  Han]{xiao2024fastcomposer}
Guangxuan Xiao, Tianwei Yin, William~T Freeman, Fr{\'e}do Durand, and Song Han.
\newblock Fastcomposer: Tuning-free multi-subject image generation with
  localized attention.
\newblock \emph{International Journal of Computer Vision}, pages 1--20, 2024.

\bibitem[Yan et~al.(2023)Yan, Zhang, Wang, Zhou, Zhang, Cheng, Yu, and
  Fu]{yan2023facestudio_similar_ip-adapter}
Yuxuan Yan, Chi Zhang, Rui Wang, Yichao Zhou, Gege Zhang, Pei Cheng, Gang Yu,
  and Bin Fu.
\newblock Facestudio: Put your face everywhere in seconds.
\newblock \emph{arXiv preprint arXiv:2312.02663}, 2023.

\bibitem[Ye et~al.(2024)Ye, Zhang, Liu, Han, and Yang]{ye2024ipadapter_git}
H.~Ye, J.~Zhang, S.~Liu, X.~Han, and W.~Yang.
\newblock Ip-adapter: Text compatible image prompt adapter for text-to-image
  diffusion models.
\newblock \url{https://github.com/tencent-ailab/IP-Adapter}, 2024.
\newblock GitHub repository.

\bibitem[Ye et~al.(2023)Ye, Zhang, Liu, Han, and Yang]{ye2023ip-adapter}
Hu~Ye, Jun Zhang, Sibo Liu, Xiao Han, and Wei Yang.
\newblock Ip-adapter: Text compatible image prompt adapter for text-to-image
  diffusion models.
\newblock \emph{arXiv preprint arXiv:2308.06721}, 2023.

\bibitem[Zhang et~al.(2023)Zhang, Rao, and
  Agrawala]{zhang2023adding_controlnet}
Lvmin Zhang, Anyi Rao, and Maneesh Agrawala.
\newblock Adding conditional control to text-to-image diffusion models.
\newblock In \emph{Proceedings of the IEEE/CVF international conference on
  computer vision}, pages 3836--3847, 2023.

\bibitem[Zhang et~al.(2018)Zhang, Isola, Efros, Shechtman, and
  Wang]{zhang2018unreasonable_LPIPS}
Richard Zhang, Phillip Isola, Alexei~A Efros, Eli Shechtman, and Oliver Wang.
\newblock The unreasonable effectiveness of deep features as a perceptual
  metric.
\newblock In \emph{Proceedings of the IEEE conference on computer vision and
  pattern recognition}, pages 586--595, 2018.

\bibitem[Zhang et~al.(2024)Zhang, Huang, Chen, Zhang, Wu, Feng, Wang, Shen,
  Liu, and Luo]{zhang2024flashface}
Shilong Zhang, Lianghua Huang, Xi~Chen, Yifei Zhang, Zhi-Fan Wu, Yutong Feng,
  Wei Wang, Yujun Shen, Yu~Liu, and Ping Luo.
\newblock Flashface: Human image personalization with high-fidelity identity
  preservation.
\newblock \emph{arXiv preprint arXiv:2403.17008}, 2024.

\bibitem[Zhao et~al.(2023)Zhao, Chen, Chen, Bao, Hao, Yuan, and
  Wong]{zhao2023uni-controlnet}
Shihao Zhao, Dongdong Chen, Yen-Chun Chen, Jianmin Bao, Shaozhe Hao, Lu~Yuan,
  and Kwan-Yee~K. Wong.
\newblock Uni-controlnet: All-in-one control to text-to-image diffusion models.
\newblock \emph{Advances in Neural Information Processing Systems}, 2023.

\bibitem[Zhou et~al.(2023)Zhou, Zhang, Sun, and
  Xu]{zhou2023enhancing_profusion_personalized}
Yufan Zhou, Ruiyi Zhang, Tong Sun, and Jinhui Xu.
\newblock Enhancing detail preservation for customized text-to-image
  generation: A regularization-free approach.
\newblock \emph{arXiv preprint arXiv:2305.13579}, 2023.

\end{thebibliography}

\end{document}